\documentclass[journal]{IEEEtran}
\usepackage[dvipsnames]{xcolor}
\usepackage{amsmath,amsfonts,graphicx,float}
\usepackage{xcolor}
\usepackage{multicol}
\usepackage{array}
\usepackage{cleveref}
\usepackage{booktabs}
\usepackage{ragged2e}
\usepackage{multicol}
\usepackage{multirow,bigstrut}
\usepackage{color}
\usepackage{tikz}
\usetikzlibrary{shapes,arrows}
\usepackage{caption}
\usepackage{xspace}
\usepackage{graphicx}
\usepackage[justification=centering]{caption}
\usepackage{subcaption}
\usepackage{lipsum}
\usepackage{cite}
\usepackage[colorlinks = fasle, linkcolor = blue, urlcolor  = blue, citecolor = green, anchorcolor = blue]{hyperref}
\usepackage{algorithm2e}
\usetikzlibrary{shapes,fit}
\usepackage{multirow}
\usepackage[pscoord]{eso-pic}
\usetikzlibrary{intersections}

\newcolumntype{L}{>{\centering\arraybackslash}m{0.035\textwidth}}
\newcolumntype{Q}{>{\centering\arraybackslash}m{0.048\textwidth}}
\newcolumntype{M}{>{\centering\arraybackslash}m{0.100\textwidth}}
\newcolumntype{N}{>{\centering\arraybackslash}m{0.050\textwidth}}
\newcolumntype{S}{>{\centering\arraybackslash}m{0.075\textwidth}}
\newcolumntype{K}{>{\centering\arraybackslash}m{0.15\textwidth}}
\newcolumntype{O}{>{\centering\arraybackslash}m{0.065\textwidth}}
\newcolumntype{J}{>{\centering\arraybackslash}m{0.2\textwidth}}
\newcolumntype{F}{>{\centering\arraybackslash}m{0.045\textwidth}}

\begin{document}
\title{No-Reference Stereoscopic Video Quality Assessment Using Joint Motion and Depth Statistics}%A Joint Model for Depth and Motion in Natural Stereoscopic Videos and its Applications to No-Reference Video Quality Assessment

\author{Balasubramanyam~Appina, Akshith~Jalli, Shanmukha~Srinivas~Battula, Sumohana~S.~Channappayya
\thanks{The authors are with the Lab for Video and Image Analysis (LFOVIA), Department
of Electrical Engineering, Indian Institute of Technology Hyderabad, Kandi,
India, 502285 e-mail: \{ee13m14p100001,~ee13b1013,~ee13b1006,~sumohana\}@iith.ac.in.}}
\date{\today}
\maketitle

\begin{abstract}
We present a no-reference (NR) quality assessment algorithm for assessing the perceptual quality of natural stereoscopic 3D (S3D) videos. This work is inspired by our finding that the joint statistics of the subband coefficients of motion (optical flow or motion vector magnitude) and depth (disparity map) of natural S3D videos possess a unique signature. Specifically, we empirically show that the joint statistics of the motion and depth subband coefficients of S3D video frames can be modeled accurately using a Bivariate Generalized Gaussian Distribution (BGGD). We then demonstrate that the parameters of the BGGD model possess the ability to discern quality variations in S3D videos. Therefore, the BGGD model parameters are employed as motion and depth quality features. In addition to these features, we rely on a frame-level spatial quality feature that is computed using a robust off-the-shelf NR image quality assessment (IQA) algorithm. These frame-level motion, depth and spatial features are consolidated and used with the corresponding S3D video's difference mean opinion score (DMOS) labels for supervised learning using support vector regression (SVR). The overall quality of an S3D video is computed by averaging the frame-level quality predictions of the constituent video frames. The proposed algorithm, dubbed Video QUality Evaluation using MOtion and DEpth Statistics (VQUEMODES) is shown to outperform the state-of-the-art methods when evaluated over the IRCCYN and LFOVIA S3D subjective quality assessment databases.
\end{abstract}


\IEEEpeerreviewmaketitle

\section{Introduction}
The advancements in digital multimedia technologies over the past decade have received tremendous consumer acceptance. Stereoscopic 3D (S3D) digital technology has received a lot of attention lately and several industries such as film, gaming, education etc., are focused on S3D content creation. 
%According to $\bf{eMarketer}$ \cite{website:emarketer}, the number of digital video viewers increased to 210 million in United States (US) in 2015, a rise of 5\% compared to 2014. 
The Motion Pictures Association of America $\bf{MPAA}$ \cite{mpaa2015} and $\bf{Statista}$ \cite{website:3dstatista} reported that in 2015, the S3D movie box office profit reached \$1.7 billion and this was an increase of 20\% compared to 2014. They also reported that the number of S3D screens in 2015 increased by 15\% compared to 2014 in the United States (US). These numbers are a clear indication of the advancements in S3D technology and its ever increasing popularity with consumers.

S3D video content creation \cite{3DChain} involves several processing steps like sampling, quantization, synthesis etc. Each step in this processing chain affects the perceptual quality of the source video content leading to a degradation of the Quality of Experience (QoE) of the end user. To guarantee high user satisfaction, it becomes important to assess the loss in quality at each step of the lossy process. Quality assessment (QA) is the process of judging or estimating the quality of a video according to its perceptual experience. QA can be categorized into two types: subjective and objective. In subjective assessment, the viewers quantify their opinion on video quality based on their perceptual feel after viewing the content. Objective quality assessment refers to the automatic quality evaluation process designed to mimic subjective assessment. It has three flavors: full reference (FR) where the pristine source is available for QA, reduced reference (RR) where partial information about the pristine source is available for QA and no-reference (NR), where no pristine source information is available for QA. In this work, we focus on S3D NR video quality assessment (VQA). 

The predominant approach to objectively assessing the quality of S3D content (image and video) has been to leverage the strength of 2D QA algorithms in conjunction with a factor that accounts for depth quality (for e.g.,\cite{yasakethu2009analyzing,hewage2009depth,regis2013objective,bosc2011towards,banitalebi2016efficient,battisti2015perceptual, Joveluro2010,Flosim3D2017}). The reliance on 2D QA metrics for S3D quality assessment could be explained by the fact that depth perception is a consequence of the relative spatial shift present in two 2D images. Another reason for relying on 2D QA methods could be the fact that they are far more mature given the ubiquity of 2D content. Further, access to publicly available S3D subjective quality databases is very limited in general, and even more so in the case of videos. In contrast to the 2D QA reliant approach, we propose an NR S3D VQA algorithm based on statistical properties that are {\em{innately stereoscopic}} and lightly augment it with a 2D NR spatial score. Towards this end, we propose a joint statistical model of the multi scale subband coefficients of motion and depth components of a natural S3D video sequence. Specifically, we propose a Bivariate Generalized Gaussian Distribution (BGGD) model for this joint distribution. We show that the BGGD model parameters possess good distortion discrimination properties. The model parameters combined with frame-wise 2D NR spatial quality scores are used as features to train an SVR for the NR VQA task. 

The rest of the paper is organized as follows: Section \ref{sec:literature survey} reviews relevant literature. Section \ref{Sec:Proposed Method} presents the proposed joint statistical model and the NR S3D VQA algorithm. Section \ref{sec:results} presents and discusses the results of the proposed NR VQA algorithm. Concluding remarks are made in Section \ref{sec:conclusions}.
\section{Background}
\label{sec:literature survey}
Given that our contribution spans S3D video modeling and VQA, we review relevant literature in appropriately titled subsections in the following.
%We first present the literature survey on experimental studies of joint dependencies between motion and depth components, and studies of S3D scene statistical modeling. Also, we review the relevant objective quality assessment models of stereoscopic image and videos.
\subsection{Motion and Depth Dependencies in the Human Visual System (HVS)}
%Young \textit{et al.} \cite{young1987gaussian} performed a series of experiments to model the receptive fields of the striate cortex. They concluded that the gaussian function is a good model of simple cells in primary visual cortex. 
Maunsell and Van Essen \cite{maunsell1983functional} performed psychovisual experiments on the monkey visual cortex to explore the disparity selectivity in the middle temporal (MT) visual area. They found that two-thirds of MT area neurons are highly tuned and responsible for binocular disparity processing. Roy \textit{et al.} \cite{roy1992disparity} experimented on a large number of MT area neurons (295 neurons) of awake monkeys to explore the dependencies between motion and depth components. They concluded that 95\% of the tested neurons were highly responsive to the crossed and uncrossed disparities. De Angelis and Newsome \cite{deangelis1999organization} performed psychovisual experiments to explore the depth tuning in the MT area. They concluded that the MT neurons are primarily responsible for motion perception and also that these neurons are important for depth perception.
Dayan \textit{et al.} \cite{dayan2003theoretical} performed an experiment on the MT area of Macaque monkeys to model the firing response rate of a neuron. They hypothesized that the frequency response of neuron firing rate can be accurately modeled with a generalized gaussian distribution. These findings combined with the observation that neuronal responses are tuned to the statistical properties of the sensory input or stimuli \cite{simoncelli2001natural} motivate us to study and model the joint statistical relationship between the subband coefficients of motion and depth components of the {\em{stimuli}} i.e., natural S3D videos. 
\subsection{S3D Natural Scene Statistical Modeling}
We briefly review literature on S3D natural scene statistical modeling. 
Huang \textit{et al.} \cite{huang2000statistics} explored the range statistics of a natural S3D scene. The range maps are captured using laser range finder and pixel, gradient and wavelet statistics are modeled. Liu \textit{et al.} \cite{liu2008disparity,liu2010dichotomy} computed the disparity information from the depth maps which were created using range finders. They established a spherically modeled eye structure to construct the disparity maps and finally correlated the constructed disparity maps with stereopsis of HVS maps. Potetz and Lee \cite{potetz2003statistical} and Liu \textit{et al.} \cite{liu2011statistical,liu2009luminance} studied the statistical relationship between the luminance and spatial disparity maps in a multiscale subband decomposition domain. They concluded that the histograms of luminance and range/disparity subband coefficients have sharp peaks and heavy tails and can be modeled with a univariate GGD (UGGD). Further, they explored the correlation dependencies between these subband coefficients. Su \textit{et al.} \cite{su2011natural} performed a study to explore the relationship between chrominance and range components. They found that the conditional distribution of chrominance coefficients (given range gradients) is modeled well using a Weibull distribution. While these studies focused on static S3D natural scenes, we found few studies on the joint statistics of motion and depth components of S3D natural videos. This paucity has provided additional motivation for our work. 
\subsection{S3D Video Quality Assessment}
We now review S3D video quality assessment methods in the following. These methods could broadly be classified into statistical modeling based and human visual system (HVS) based approaches. 

Statistical model based approaches have been very successful in S3D IQA \cite{chen2013no,khan2015full,su2015oriented,appina2016no,hachicha2017no}. Mittal \textit{et al.} \cite{mittal2011algorithmic} proposed an NR S3D VQA metric based on statistical measurements of disparity and differential disparity. They computed the mean, median, kurtosis, skew-ness and standard deviation values of the disparity and differential disparity maps to estimate the quality of an S3D video. These approaches point to the efficacy of using statistical models for S3D QA, and provide the grounding for our work.   

As discussed earlier, several objective S3D FR VQA models \cite{yasakethu2009analyzing,hewage2009depth,regis2013objective,Joveluro2010,bosc2011towards,banitalebi2016efficient,battisti2015perceptual} are based on applying 2D IQA/VQA algorithms on individual views and the depth component of an S3D video. In addition to these, we review a few HVS inspired FR, RR and NR VQA algorithms next. \cite{jin2011frqa3d} and \cite{Han2012VCIP} proposed S3D VQA models based on 3D-Discrte Cosine Transform (DCT) and 3D structer models from the spatial, temporal and depth components of an S3D view. Qi \textit{et al.} \cite{qi2016stereoscopic} proposed an FR S3D VQA metric based on measuring the Just Noticeable Differences (JND) on spatial, temporal and depth maps. They computed the temporal JNDs (TJND) of spatial component and interview JNDs (IJND) from TJNDs to estimate the binocular property. Further, they calculated the similarity maps between reference and distorted spatial JNDs to estimate the spatial quality and finally, they computed the mean across all JNDs to estimate the overall S3D video quality score. De Silva \textit{et al.} \cite{de20103d} proposed an FR S3D VQA metric based on measuring the perceivable distortion strength from depth maps. They computed the JND value between reference and distorted depth maps to measure the quality of an S3D video. Galkandage \textit{et al.} \cite{galkandage2016stereoscopic} proposed S3D FR IQA and VQA metrics based on an HVS model and temporal features. They computed the Energy Quality Metric (EBEQM) scores to measure the spatial quality and finally pooled these scores by using empirical methods to estimate the overall quality score of an S3D video. De Silva \textit{et al.} \cite{de2013toward} proposed an S3D FR VQA based on measuring the spatial distortion, blur measurement and content complexity. They measured the structural similarity and edge degradation between reference and distorted views to compute the spatial quality, distortion strength. Content complexity was measured by calculating the spatial and temporal indices of an S3D view.

Hewage and Martini \textit{et al.} \cite{hewage2011reduced} proposed an S3D RR VQA metric based on depth map edges and S3D view chrominance information. They applied the Sobel operator to compute the edges of a depth map and utilized these edges to extract the chrominance features of an S3D view. Finally, they computed the PSNR values of extracted features to estimate the quality of an S3D video.
Yu \textit{et al.} \cite{yu2016binocular} proposed an S3D RR VQA metric based on perceptual properties of the HVS. They relied on motion vector strength to predict the reduced reference frame of a reference video, and binocular fusion and rivalry scores were calculated using the RR frames. Finally these scores were pooled using motion intensities as weights to compute the quality score of an S3D video. 

Sazzad \textit{et al.} \cite{sazzad2010spatio} proposed an S3D NR VQA metric based on spatiotemporal segmentation. They measured structural loss by computing the edge strength degradation in each segment and motion vector length was measured to estimate the temporal cue loss. 
Ha and Kim \cite{ha2011perceptual} proposed an S3D NR VQA metric based on temporal variance, intra and inter disparity measurements. Depth maps are computed by minimizing the MSE values, and motion vector length is calculated to estimate the temporal variations. Intra and inter frame disparities were computed to measure the dependencies between motion and depth components. Solh and AlRegib \cite{solh2011no} proposed an S3D NR VQA metric based on temporal inconsistencies, spatial and temporal outliers. Spatial and temporal outliers were measured by calculating the difference between ideal and estimated depth maps, and temporal inconsistencies were computed by calculating the standard deviation of difference between depth maps of successive frames. 
Hasan \textit{et al.} \cite{hasan2014no} proposed an S3D NR VQA metric based on similarity matches and edge visualized areas. They utilized the edge strength to find the visualized areas and computed the energy error of similarity measure estimated between left and right views to calculate the disparity index. 
Silva \textit{et al.} \cite{silva2015no} proposed an S3D NR VQA metric based on distortion strength, disparity and temporal depth qualities of a 3D video. They computed the depth cue loss by measuring the spatial distortion strength, and temporal depth qualities were measured by calculating the correlation between histograms of frame motion vectors. 
Han \textit{et al.} \cite{han2015extended} proposed an NR S3D VQA metric based on the encoder settings of a transmitted video. They used the ITU-T G.1070 settings to model the packet loss artefacts and quality was measured by computing the correlation between perceptual scores and packet loss rates at different bit rates.  Mahamood and Ghani \cite{mahmood2015objective} proposed an S3D NR VQA metric based on computing the motion vector lengths and depth map features. They concluded that the number of bad frames in a video is a good predictor of motion and depth quality of an S3D video. 
Yang \textit{et al.} \cite{yang2017} proposed an S3D NR VQA metric based on multi view binocular perception model. They applied the curvelet transform on spatial information of an S3D video to extract the texture analysis features and optical flow features were utilized to measure the temporal quality. Finally, they used empirical weight combinations to pool these scores to compute the overall quality score. Chen \textit{et al.} \cite{chen2017blind} proposed an S3D NR VQA model based on binocular energy mechanism. They computed the auto-regressive prediction based disparity measurement and natural scene statistics of an S3D video to compute the quality.

%%%%%%%%%%%%%%%%%%%%%%%%%%%%%%%%%%
\begin{figure*}
\captionsetup[subfigure]{justification=centering}
\centering
\begin{subfigure}[b]{0.32\textwidth}
\includegraphics[width=5.5cm,height=3cm]{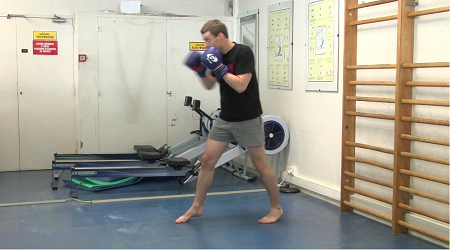} % [width=5cm,height=3cm]
\subcaption{\small Reference left view first frame.}
\label{fig:l1}
\end{subfigure}
\begin{subfigure}[b]{0.32\textwidth}
\includegraphics[width=5.5cm,height=3cm]{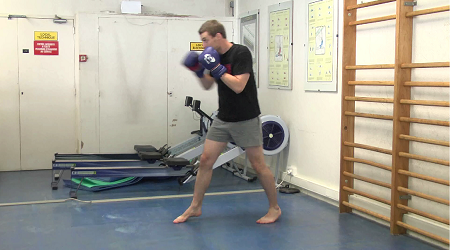}
\subcaption{\small Reference right view first frame.}
\label{fig:r1} 
\end{subfigure}
\begin{subfigure}[b]{0.32\textwidth}
\includegraphics[width=5.5cm,height=3cm]{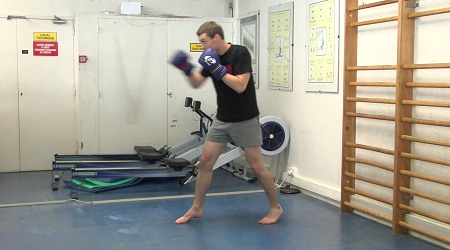}
\subcaption{\small Reference left view second frame.} 
\label{fig:l2}
\end{subfigure}
\\
\begin{subfigure}[b]{0.32\textwidth}
\includegraphics[width=5.5cm,height=3cm]{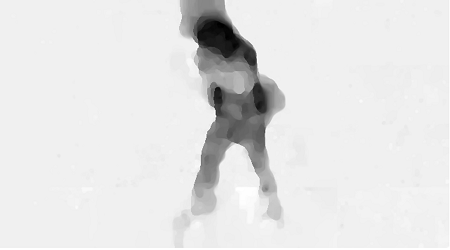}
\subcaption{\small Optical flow magnitude map.} 
\label{fig:motionr}
\end{subfigure}
\begin{subfigure}[b]{0.32\textwidth}
\includegraphics[width=5.5cm,height=3cm]{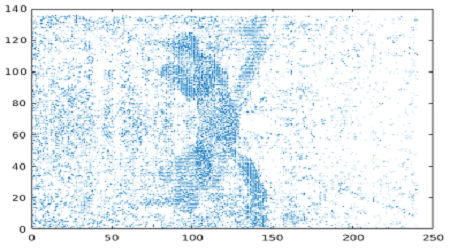}
\subcaption{\small Motion vector magnitude map.} 
\label{fig:mr}
\end{subfigure}
\begin{subfigure}[b]{0.32\textwidth}
\includegraphics[width=5.5cm,height=3cm]{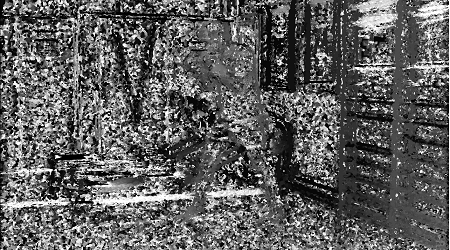}
\subcaption{\small Disparity map.} 
\label{fig:depthr}
\end{subfigure}
\\
\begin{subfigure}[b]{0.32\textwidth}
\includegraphics[width=5.5cm,height=3cm]{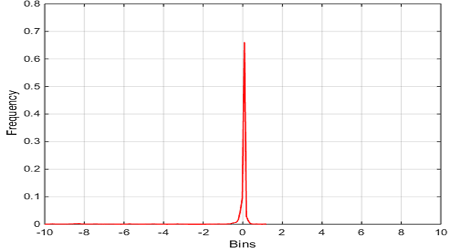}
\subcaption{\small Optical flow magnitude histogram.} 
\label{fig:motionhist}
\end{subfigure}
\begin{subfigure}[b]{0.32\textwidth}
\includegraphics[width=5.5cm,height=3cm]{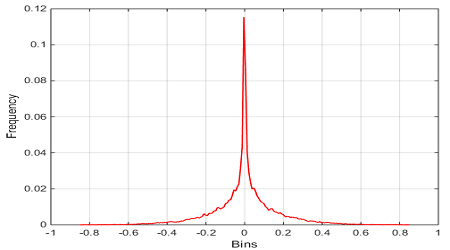}
\subcaption{\small Motion vector magnitude histogram.} 
\label{fig:motionhistm}
\end{subfigure}
\begin{subfigure}[b]{0.32\textwidth}
\includegraphics[width=5.5cm,height=3cm]{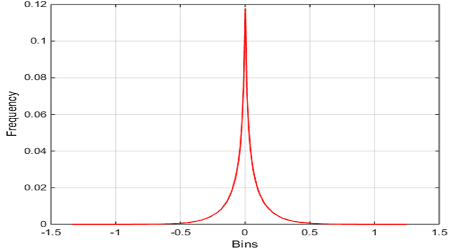}
\subcaption{\small Disparity map histogram.} 
\label{fig:depthhist}
\end{subfigure}
\\
\begin{subfigure}[b]{0.45\textwidth}
\includegraphics[height=4.5cm]{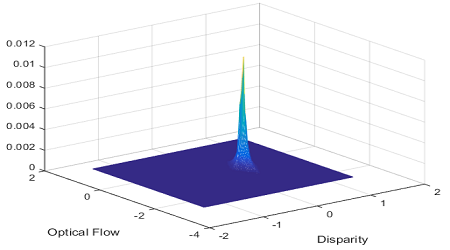}
\subcaption{\small Joint histogram plot of optical flow magnitude and disparity maps.} 
\label{fig:3dhist}
\end{subfigure}
\begin{subfigure}[b]{0.5\textwidth}
\includegraphics[height=4.5cm]{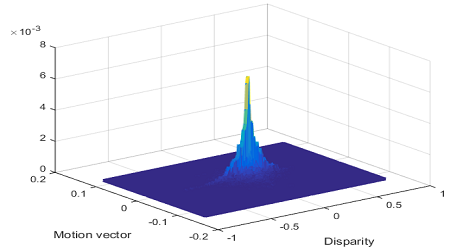}
\subcaption{\small Joint histogram plot of motion vector magnitude and disparity maps.}
\label{fig:3dhistm} 
\end{subfigure}
\\
\begin{subfigure}[b]{0.45\textwidth}
\includegraphics[height=4.5cm]{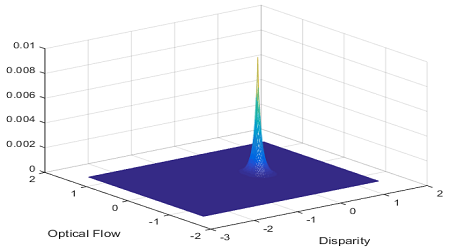}
\subcaption{\small Bivariate GGD fit between optical flow magnitude and disparity maps.} 
\label{fig:3dmodelhist}
\end{subfigure}
\begin{subfigure}[b]{0.5\textwidth}
\includegraphics[height=4.5cm]{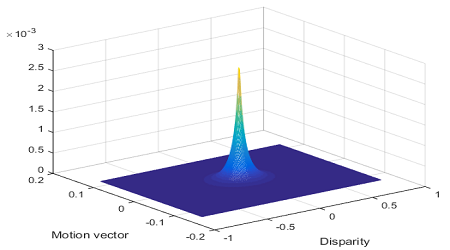}
\subcaption{\small Bivariate GGD fit between motion vector magnitude and disparity maps.} 
\label{fig:3dmodelhistm}
\end{subfigure}
\caption{Illustration of BGGD model fit between combinations of optical flow magnitude and disparity maps, motion vector magnitude and disparity maps.}
\label{fig:LumDepth}
\end{figure*}
%%%%%%%%%%%%%%%%%%%%%%%%%%%%%%%%%% 
Our literature survey has equipped us with the required background and motivation to study and model the joint statistics of motion and depth in S3D natural videos in a multi-resolution analysis domain. Further, it has given us the grounding to propose an S3D NR VQA algorithm dubbed Video QUality Evaluation using MOtion and DEpth Statistics (VQUEMODES) that relies on the joint statistical model parameters and 2D NR IQA scores. We describe the proposed approach in detail in the following section.

\section{Proposed Method} 
\label{Sec:Proposed Method}
We first analyze the joint statistical behavior of the subband coefficients of motion and depth components in S3D natural videos. We then propose a BGGD model for the joint distribution. Subsequently, we describe the proposed S3D NR VQA algorithm. 
\subsection{BGGD Modeling}
We empirically show that a Bivariate Generalized Gaussian Distribution (BGGD) accurately captures the dependencies between motion and depth subband coefficients. We use both optical flow vector magnitude and motion vector magnitude to represent motion in our statistical analysis. We do so to investigate the behavior at both fine and coarse motion representations respectively. Further, motion vector computation has significantly lower computational complexity compared to optical flow computation. We perform our analysis using multi-scale (3 scales) and multi-orientation ($0^0,30^0,60^0,90^0,120^0,150^0$) subband decomposition coefficients of the motion (optical flow/motion vector) and disparity maps. 

The multivariate GGD distribution of a random vector ${\bf{x}} \in \mathbb{R}^N$ is given by \cite{pascal2013parameter}
\begin{align}
p({\bf{x}}|{\bf{M}},\alpha,\beta)&=\frac{1}{|{\bf{M}}|^\frac{1}{2}}g_{\alpha,\beta}({\bf{x}}^T{\bf{M}}^{-1}{\bf{x}}) ,\\
g_{\alpha,\beta}(y)&=\frac{\beta\Gamma(\frac{N}{2})}{(2^{\frac{1}{\beta}}\Pi\alpha)^{\frac{N}{2}}\Gamma(\frac{N}{2\beta})}e^{-\frac{1}{2}(\frac{y}{\alpha})^{\beta}},
\label{eqn:mggd}
\end{align}
where ${\bf{M}}$ is an $N\times N$ symmetric scatter matrix, $\beta$ is the shape parameter, $\alpha$ is the scale parameter and $g_{\alpha, \beta}(\cdot)$ is the density generator. 

Figs. \ref{fig:l1} and \ref{fig:l2} show the first and second frames of the Boxers S3D video left view respectively, and Fig. \ref{fig:r1} shows the first frame of the right view from the Boxers video sequence of the IRCCYN database \cite{urvoy2012nama3ds1}. Fig. \ref{fig:depthr} shows the disparity map estimated using the SSIM-based stereo matching algorithm \cite{chen2013full} computed between first frame of left and right views. Figs. \ref{fig:motionr} and \ref{fig:mr} show the optical flow and motion vector maps computed between the first and second frames of the left view respectively. The optical flow map is computed using the Black and Anandan \cite{black1993framework} flow algorithm and the motion vector map is estimated using the three-step block motion estimation algorithm \cite{jakubowski2013block}. Figs. \ref{fig:depthhist}, \ref{fig:motionhist} and \ref{fig:motionhistm} show the histograms of the subband coefficients of disparity, optical flow magnitude and motion vector magnitude respectively. Fig. \ref{fig:3dhist} shows the joint histogram between disparity and optical flow magnitude subband coefficients, and Fig. \ref{fig:3dmodelhist} shows the estimated BGGD model of \ref{fig:3dhist}. Fig. \ref{fig:3dhistm} shows the joint histogram of disparity and motion vector magnitude subband coefficients and Fig. \ref{fig:3dmodelhistm} shows the estimated BGGD model of \ref{fig:3dhistm}. The BGGD model parameters were estimated using the approach taken by Su \textit{et al.} \cite{su2015oriented}. The marginal and joint histogram plots were computed at the first scale and $0^{0}$ orientation of the steerable pyramid decomposition. From the histograms \ref{fig:depthhist}, \ref{fig:motionhist} and \ref{fig:motionhistm} we see that the subband coefficients have sharp peaks and long tails and is consistent with the observations in \cite{dayan2003theoretical,young1987gaussian}. The joint histograms between disparity and motion components \ref{fig:3dmodelhist} and \ref{fig:3dmodelhistm} also have sharp peaks and heavy tails. These empirical findings lead us to propose that a bivariate GGD ($N = 2$) is ideally suited to model the joint distribution of disparity and motion subband coefficients.

The efficacy of the proposed BGGD model is first evaluated on three publicly available S3D video databases: IRCCYN \cite{urvoy2012nama3ds1}, RMIT \cite{cheng2012rmit3dv} and LFOVIA \cite{appina2016subjective1}. The IRCCYN video database has pristine S3D videos and their H.264 and JP2K distorted versions, the RMIT database comprises pristine S3D videos and the LFOVIA database is composed of pristine S3D videos and their H.264 compressed versions.
\begin{figure*}[htbp]
\centering
\begin{subfigure}[b]{0.32\textwidth}
\includegraphics[height=3cm]{Figures/l1.png}% [width=5cm,height=3cm] width=5.5cm,
\subcaption{\small Reference left view.}
\label{fig:1l1}
\end{subfigure}
\begin{subfigure}[b]{0.32\textwidth}
\includegraphics[height=3cm]{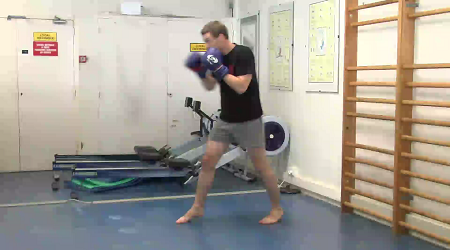}
\subcaption{\small H.264 compressed left view.} 
\label{fig:1ld}
\end{subfigure}
\begin{subfigure}[b]{0.32\textwidth}
\includegraphics[height=3cm]{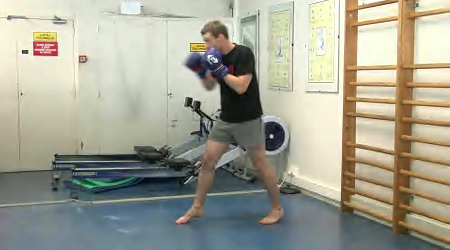}
\subcaption{\small JP2K distorted left view.} 
\label{fig:1ld1}
\end{subfigure}
\\
\begin{subfigure}[b]{0.32\textwidth}
\includegraphics[height=3cm]{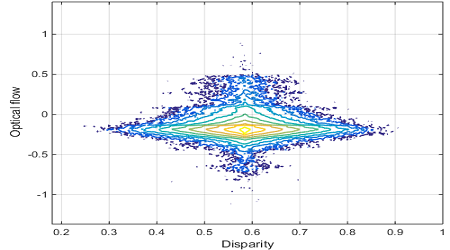}
\subcaption{\small Isoprobability contour plot of joint optical flow magnitude and disparity distribution of the reference view. Best fit BGGD model: $\alpha_o = 5\times 10^{-14}$, $\beta_o = 0.0405$, $\chi_o = 1\times10^{-7}$.}
\label{fig:13dcontref}
\end{subfigure}
\begin{subfigure}[b]{0.32\textwidth}
\includegraphics[height=3cm]{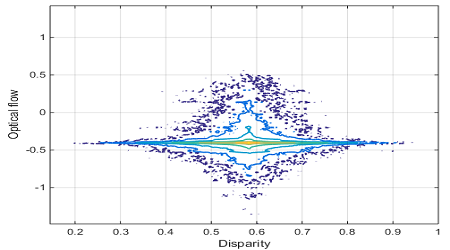}
\subcaption{\small Isoprobability contour plot of joint optical flow magnitude and disparity distribution of the H.264 distorted view. Best fit BGGD model: $\alpha_o = 1\times10^{-8} $, $\beta_o = 0.0514$, $\chi_o = 1\times10^{-8}$.} 
\label{fig:13dconth264}
\end{subfigure} 
\begin{subfigure}[b]{0.32\textwidth}
\includegraphics[height=3cm]{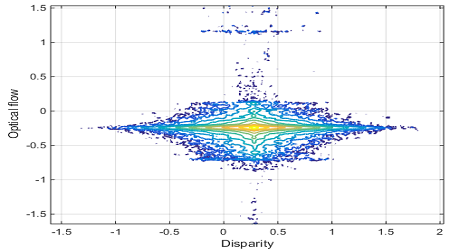}
\subcaption{\small Isoprobability contour plot of joint optical flow magnitude and disparity distribution of the JP2K distorted view. Best fit BGGD model: $\alpha_o = 1\times10^{-16} $, $\beta_o = 0.4238$, $\chi_o = 5\times10^{-7}$.}  
\label{fig:13dcontjp}
\end{subfigure}
\\
\begin{subfigure}[b]{0.32\textwidth}
\includegraphics[height=3cm]{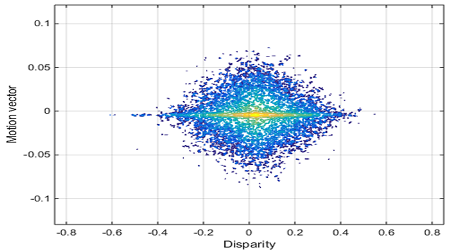}
\subcaption{\small Isoprobability contour plot of joint motion vector magnitude and disparity distribution of reference view. Best fit BGGD model: $\alpha_m = 4\times10^{-5}$, $\beta_m = 0.3579$, $\chi_m = 1 \times 10^{-8}$.}
\label{fig:13dcontmref}
\end{subfigure}
\begin{subfigure}[b]{0.32\textwidth}
\includegraphics[height=3cm]{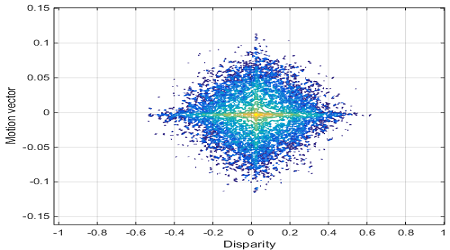}
\subcaption{\small Isoprobability contour plot of joint motion vector magnitude and disparity distribution of H.264 distorted view. Best fit BGGD model: $\alpha_m = 4\times10^{-4} $, $\beta_m = 0.4457 $, $\chi_m =8\times10^{-7} $.} 
\label{fig:13dcontmh264}
\end{subfigure}
\begin{subfigure}[b]{0.32\textwidth}
\includegraphics[height=3cm]{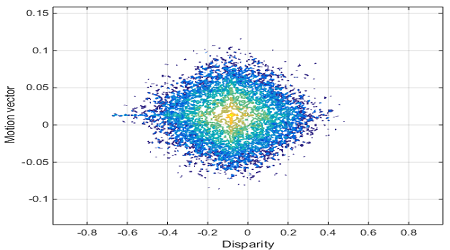}
\subcaption{\small Isoprobability contour plot of joint motion vector magnitude and disparity distribution of JP2K distorted view. Best fit BGGD model: $\alpha_m = 6\times10^{-7}$, $\beta_m = 0.603$, $\chi_m = 6\times10^{-7}$.}  
\label{fig:13dcontmjp}
\end{subfigure}
\caption{Effects of distortion on the joint distribution of motion and depth subband coefficients ($0^0$ orientation and first scale). It can be seen that the BGGD model parameters are able to track the effects of distortions.}
\label{fig:LumDepthDist}
\end{figure*}
%%%%%%%%%%%%%%%%%%%%%%%%%%%%%%%%
\begin{table*}[htbp]
\centering 
\caption{BGGD features ($\alpha$, $\beta$) and goodness of fit ($\chi$) value of reference S3D videos and its symmetric distortion combinations of IRCCYN, LFOVIA and RMIT databases.}
\begin{tabular}{|c|c|c|c|c|c|c|c|c|c|}
\hline
Video Frame &Distortion Type & \multicolumn{2}{c|}{Test stimuli} & \multicolumn{6}{c|}{BGGD features ($\alpha$, $\beta$) and goodness of fit ($\chi$) value}\\
\cline{3-10} 
& &  Left & Right &  $\alpha_{o}$ & $\beta_{o}$ & $\chi_{o}$ & $\alpha_{m}$ & $\beta_{m}$ & $\chi_{m}$\\
\hline
\multirow{8}{*}{\begin{minipage}{.2\textwidth}
\centering
%\vspace{-0.6cm}
\includegraphics[width=2.8cm,height=1.5cm]{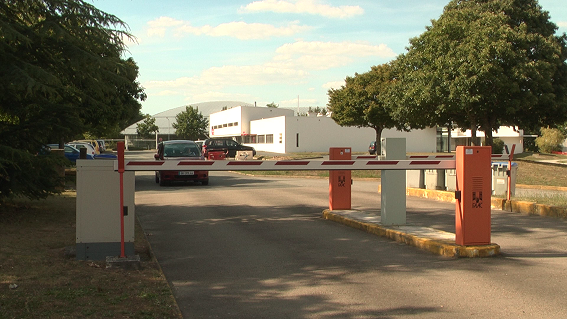}
\end{minipage}
} & Reference & - & - &  1 $\times$ $10^{-5}$&0.337 &3 $\times$ $10^{-6}$& 8 $\times$ $10^{-6}$&0.312 &	2 $\times$ $10^{-6}$ \\
\cline{2-10}
& & 32 & 32 & 2 $\times$ $10^{-6}$&	0.213 &	7 $\times$ $10^{-6}$& 6 $\times$ $10^{-7}$&	0.245 &	5 $\times$ $10^{-6}$ \\
\cline{3-10}
& H.264  &  38 &  38 &3 $\times$ $10^{-5}$&	0.341 &	3 $\times$ $10^{-6}$& 8 $\times$ $10^{-5}$&	0.377 &	2 $\times$ $10^{-6}$ \\
\cline{3-10}
&  (QP) &  44 &  44 & 1 $\times$ $10^{-4}$&	0.407 &	2 $\times$ $10^{-6}$& 7 $\times$ $10^{-4}$&	0.425 &	1 $\times$ $10^{-6}$ \\
\cline{2-10}
& & 2 & 2 & 3 $\times$ $10^{-4}$&0.715&	1 $\times$ $10^{-6}$&4 $\times$ $10^{-6}$	&0.737 &1 $\times$ $10^{-7}$\\
\cline{3-10}
&JP2K & 8 & 8 & 1 $\times$ $10^{-4}$&0.698 &2 $\times$ $10^{-7}$&3 $\times$ $10^{-5}$&0.724 &2 $\times$ $10^{-7}$\\
\cline{3-10}
&  (Bitrate = Mb/s) & 16 & 16 & 1 $\times$ $10^{-5}$&0.619 &4 $\times$ $10^{-7}$&4 $\times$ $10^{-4}$&0.641 &3 $\times$ $10^{-7}$\\
\cline{3-10}
&  & 32 &  32 & 2 $\times$ $10^{-4}$&0.455 &1 $\times$ $10^{-6}$&5 $\times$ $10^{-4}$&0.473 &1 $\times$ $10^{-6}$\\
\hline
\multirow{8}{*}{\begin{minipage}{.2\textwidth}
\centering
\vspace{-0.6cm}
\includegraphics[width=2.8cm,height=1.5cm]{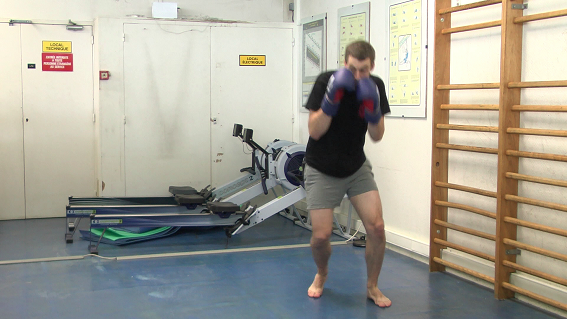}
\end{minipage}
}& Reference & - & - & 3 $\times$ $10^{-11}$&0.652 &2 $\times$ $10^{-7}$&5 $\times$ $10^{-4}$&	0.496 &	2 $\times$ $10^{-7}$ \\
\cline{2-10}
& & 32 & 32 & 6 $\times$ $10^{-6}$&	0.284 &	3 $\times$ $10^{-7}$&5 $\times$ $10^{-5}$&	0.349 &	1 $\times$ $10^{-6}$ \\
\cline{3-10}
& H.264  &  38 &  38 & 1 $\times$ $10^{-4}$&0.372 &	2 $\times$ $10^{-6}$&1 $\times$ $10^{-4}$&	0.395 &	1 $\times$ $10^{-6}$ \\
\cline{3-10}
&  (QP) &  44 &  44 & 2 $\times$ $10^{-4}$&	0.405 &	1 $\times$ $10^{-6}$&3 $\times$ $10^{-4}$&	0.441 &	9 $\times$ $10^{-7}$ \\
\cline{2-10}
& & 2 & 2 & 1 $\times$ $10^{-11}$&	0.622 &	2 $\times$ $10^{-7}$&2 $\times$ $10^{-9}$&	0.649 &	2 $\times$ $10^{-7}$ \\
\cline{3-10}
&JP2K & 8 & 8 & 1 $\times$ $10^{-11}$&	0.677 &	2 $\times$ $10^{-7}$&	2 $\times$ $10^{-10}$&	0.611 &	2 $\times$ $10^{-7}$ \\
\cline{3-10}
&  (Bitrate = Mb/s) & 16 & 16 &4 $\times$ $10^{-11}$&0.585&2 $\times$ $10^{-7}$&6 $\times$ $10^{-10}$&	0.600 &	2 $\times$ $10^{-7}$ \\
\cline{3-10}
&  & 32 &  32 &2 $\times$ $10^{-8}$&0.548 &	1 $\times$ $10^{-7}$&1 $\times$ $10^{-11}$&	0.571 &	2 $\times$ $10^{-7}$ \\
\hline
\multirow{7}{*}{\begin{minipage}{.2\textwidth}
\centering
\vspace{-0.6cm}
\includegraphics[width=2.8cm,height=1.5cm]{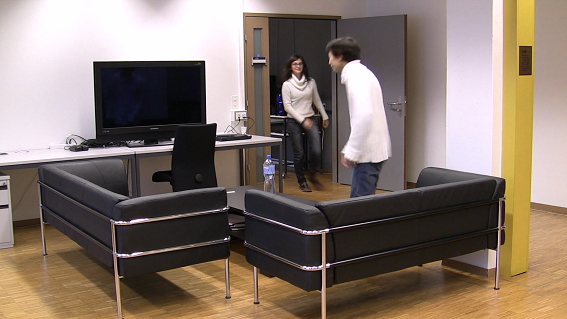}
\end{minipage}
}& Reference & - & - & 5 $\times$ $10^{-5}$&0.361 &	7 $\times$ $10^{-7}$ &5 $\times$ $10^{-5}$&	0.361 &	7 $\times$ $10^{-7}$  \\
\cline{2-10}
& & 100 & 100 & 2 $\times$ $10^{-6}$&	0.318 &	3 $\times$ $10^{-6}$ &7 $\times$ $10^{-5}$&	0.365 &	1 $\times$ $10^{-6}$  \\
\cline{3-10}
& H.264  &  200 &  200 &2 $\times$ $10^{-5}$&	0.447 &	8 $\times$ $10^{-7}$ &3 $\times$ $10^{-6}$&	0.281 &	4 $\times$ $10^{-6}$  \\
\cline{3-10}
&  (Bitrate=Kbps) & 350 & 350 & 9 $\times$ $10^{-5}$&0.435&	5 $\times$ $10^{-7}$ &2 $\times$ $10^{-6}$&	0.268 &	3 $\times$ $10^{-6}$  \\
\cline{3-10}
& & 1200 & 1200 & 2 $\times$ $10^{-4}$&	0.438 &	3 $\times$ $10^{-7}$ &2 $\times$ $10^{-5}$&	0.332 &	1 $\times$ $10^{-6}$  \\
\hline
\multirow{7}{*}{\begin{minipage}{.2\textwidth}
\centering
\vspace{-0.6cm}
\includegraphics[width=2.8cm,height=1.5cm]{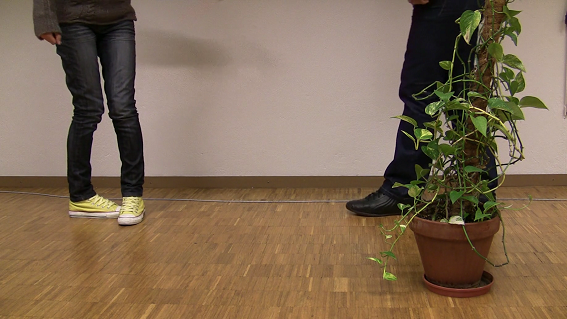}
\end{minipage}
}& Reference & - & - & 1 $\times$ $10^{-4}$&0.205 &	9 $\times$ $10^{-6}$&2 $\times$ $10^{-7}$&0.230 &5 $\times$ $10^{-6}$  \\
\cline{2-10}
& & 100 & 100 &9 $\times$ $10^{-5}$&0.430 &	3 $\times$ $10^{-6}$&4 $\times$ $10^{-4}$&	0.449 &	1 $\times$ $10^{-6}$  \\
\cline{3-10}
& H.264  &  200 &  200 & 3 $\times$ $10^{-5}$&0.288 &5 $\times$ $10^{-6}$&4 $\times$ $10^{-5}$&	0.351 &	2 $\times$ $10^{-6}$  \\
\cline{3-10}
&  (Bitrate=Kbps) & 350 & 350 &6 $\times$ $10^{-5}$&0.268 &	6 $\times$ $10^{-6}$&1 $\times$ $10^{-5}$&0.319 &4 $\times$ $10^{-6}$  \\
\cline{3-10}
& & 1200 & 1200 & 5 $\times$ $10^{-5}$&	0.240 &	1 $\times$ $10^{-6}$&3 $\times$ $10^{-6}$&0.277 &4 $\times$ $10^{-6}$  \\
\hline
\multirow{7}{*}{\begin{minipage}{.2\textwidth}
\centering
%\vspace{-0.6cm}
\includegraphics[width=2.8cm,height=1.5cm]{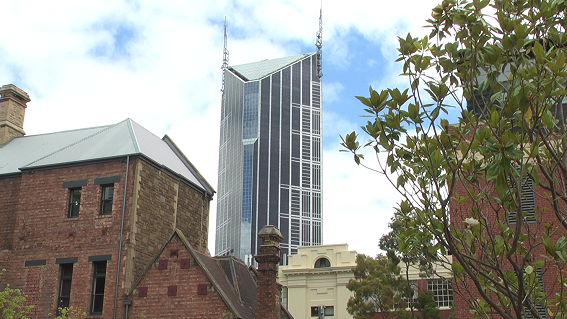}
\end{minipage}
}& &  &  &  & & &	&	 &	 \\
& Reference & - & - & 9 $\times$ $10^{-4}$ & 0.928 & 7 $\times$ $10^{-6}$ & 8 $\times$ $10^{-4}$&0.758 &8 $\times$ $10^{-7}$ \\
%\cline{3-10}
& &  &  &  & & &	&	 &	 \\
%\cline{3-10}
& &  &  &  & & &	&	 &	 \\
%\cline{3-10}
& &  &  &  & & &	&	 &	 \\
%\cline{3-10}
& &  &  &  & & &	&	 &	 \\
\hline
\multirow{7}{*}{\begin{minipage}{.2\textwidth}
\centering
%\vspace{-0.6cm}
\includegraphics[width=2.8cm,height=1.5cm]{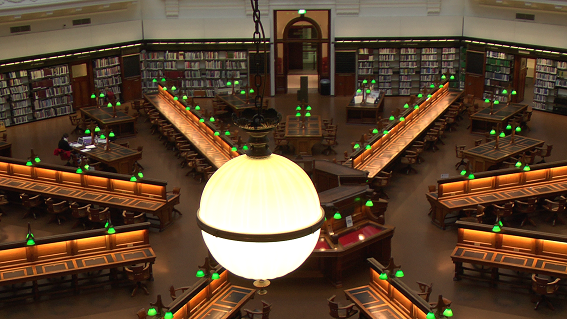}
\end{minipage}
}& &  &  &  & & &	&	 &	 \\
& Reference & - & - & 4 $\times$ $10^{-5}$ & 0.279 & 6 $\times$ $10^{-7}$&6 $\times$ $10^{-5}$&0.368 &5 $\times$ $10^{-8}$ \\
%\cline{3-10}
& &  &  &  & & &	&	 &	 \\
%\cline{3-10}
& &  &  &  & & &	&	 &	 \\
%\cline{3-10}
& &  &  &  & & &	&	 &	 \\
%\cline{3-10}
& &  &  &  & & &	&	 &	 \\
\hline
\end{tabular}
\label{tab:modelscores}
\end{table*}
Table \ref{tab:modelscores} shows the estimated BGGD model fitting parameters between disparity and motion (optical flow and motion vector) subband coefficients of the $100^{th}$ frame of a few reference S3D video sequences and their symmetric distortion combinations of the IRCCYN, LFOVIA and RMIT databases. 
The results are computed at the first scale and $0^{0}$ orientation of the steerable pyramid decomposition. ($\alpha_{o}, \beta_{o}$) correspond to optical flow parameters and ($\alpha_{m}, \beta_{m}$) correspond to motion vector parameters. $\chi$ indicates the goodness of fit value and it represents how well the proposed model fits our observations (joint histogram). The low values of the $\chi$ point to the accuracy of the proposed model. $\chi_{o}$ is the goodness of fit value computed for the optical flow case. Similarly, $\chi_{m}$ is goodness of fit value for the  motion vector case. In our analysis we observed that $\chi_o$, $\chi_m$ are in the range $10^{-8}$ and $10^{-6}$ for all S3D video sequences. From these results it is clear that the proposed BGGD model performs well in capturing the joint statistical dependencies between motion and disparity subband coefficients.

\subsection{Distortion Discrimination}
Since we are interested in NR VQA of S3D videos, we explored the efficacy of the proposed model on distorted videos as well. Fig. \ref{fig:1l1} shows the first frame of left reference view, and Figs. \ref{fig:1ld} and \ref{fig:1ld1} show the first frame of H.264 and JP2K distorted videos of corresponding reference view respectively. Due to a lack of other distortion types in the publicly available S3D video databases, we limited our analysis to H.264 and JP2K distortions. Figs. \ref{fig:13dcontref}, \ref{fig:13dconth264} and \ref{fig:13dcontjp} show the isoprobability contour plots of the joint optical flow magnitude and disparity subband coefficient distribution of the reference, H.264 and JP2K distorted frames respectively. These plots are computed at the first scale and $0^{0}$ orientation of the steerable pyramid decomposition. The plots clearly show the strong dependencies between motion and depth components of an S3D view and further, the variation in the dependencies due to distortion. As a consequence, the effects of distortion manifest themselves in a change in the BGGD model parameters. Further, it can also be seen that H.264 and JP2K distortions result in distributions with heavier tails. We observed the same trend in the joint motion vector magnitude and disparity subband coefficients distributions. 

\begin{figure}
%\center
\includegraphics[width=8cm]{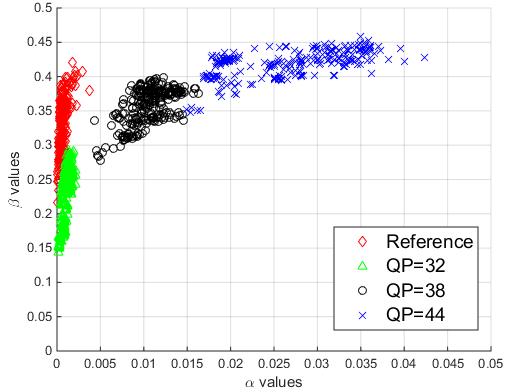}
\caption{BGGD model parameter variation of reference and H.264 distortion levels of the Boxers video sequence of IRCCYN S3D database. Each point in the scatter plot represents the model parameter pair ($\alpha_m, \beta_m$) for a frame-pair in the corresponding video. The ($\alpha_m, \beta_m$) pair clearly discriminate varying quality levels.}
\label{fig:featsetrefall}
\end{figure}

Figs. \ref{fig:13dcontmref}, \ref{fig:13dcontmh264} and \ref{fig:13dcontmjp} show the isoprobability contour plots of the joint motion vector magnitude and disparity subband coefficient distribution of the reference, H264 and JP2K distorted frames respectively. As with optical flow, these plots correspond to the subband coefficients at the first scale and $0^0$ orientation of the steerable pyramid decomposition. Again, we see a similar distortion tracking trend with the corresponding BGGD model parameters. 
%%%%%%%%%%%%%%%%%%%%%%%%%%%
%The IRCCYN database is a combination of H.264 and JP2K distortions, the LFOVIA database has H.264 distorted S3D videos. The RMIT database has only pristine sequences and it does not have any distorted videos. 
Fig. \ref{fig:featsetrefall} shows the distribution of BGGD coefficients ($\alpha_m$, $\beta_m$) computed at the first scale and $0^{0}$ orientation of all frames of the reference Boxers video and its H.264 distorted sequences at different QP (QP = 32, 38, 44) levels. It is clear that the BGGD features are able to discriminate the quality variations in S3D videos. As mentioned earlier, we consider motion vector magnitude in our work primarily due to their amenability for fast implementations. From Figs. \ref{fig:LumDepthDist} and \ref{fig:featsetrefall}, we have demonstrated that they have very good distortion discrimination properties as well.
%%%%%%%%%%%%%%%%%%%%%%%%%%%%%%%%%%%%%%%%%%%

\subsection{Video Quality Algorithm}
\begin{figure}
\center
\begin{center}
\tikzstyle{block} = [rectangle, draw, fill=white!100, text width=7em, text centered, minimum height=2em]
\tikzstyle{block1} = [rectangle, draw, fill=white!100, text width=9em, text centered, minimum height=2em]
\tikzstyle{block2} = [rectangle, draw, fill=white!100, text width=13em, text centered, minimum height=2em]
\tikzstyle{block3} = [rectangle, draw, fill=white!100, text width=17em, text centered, minimum height=2em]
\tikzstyle{line} = [draw, color=black!200, line width=0.5mm,-latex']
\tikzstyle{line1} = [draw, color=black!200, -latex']
\vspace{1cm}
%%%	
\begin{tikzpicture}[ node distance = 2.5cm,auto]
	    % Place nodes
	   
\node [block,draw=orange,line width=0.5mm] (A) {\small Stereoscopic video};
\node [block, draw=magenta, below of=A, node distance = 1.5cm, xshift=-2cm] (B) {\small Left video};
\node [block, draw=magenta, below of=A, node distance = 1.5cm, xshift=2cm] (C) {\small Right video};
\node[rectangle,draw=blue,line width=0.5mm,dashed, fit=(A) (B) (C)](D) {};

\node [block1, draw=White, below of=D, node distance = 2.5cm, xshift=-2cm] (E) {\bf{Temporal Features}};
\node [block1, draw=Violet, below of=E, node distance = 1cm, xshift=+0.3cm] (I) {\small Motion Vector Estimation};
\node [block1, draw=Violet, below of=I, node distance = 1.5cm] (J) {\small Steerable Pyramid Decomposition};
\node[rectangle,draw=green,line width=0.5mm,dashed, fit=(I) (J)](K) {};

\node [block1, draw=White, below of=D, node distance = 2.5cm, xshift=2.2cm] (M) {\bf{Depth Features}};
\node [block1, draw=Violet, below of=M, node distance = 1cm, xshift=0cm] (N) {\small Depth Map Estimation};
\node [block1, draw=Violet, below of=N, node distance = 1.5cm] (O) {\small Steerable Pyramid Decomposition};
\node[rectangle,draw=VioletRed,line width=0.5mm,dashed, fit=(N) (O)](P) {};
%
%%%
\node [block1, draw=White, below of=D, node distance = 8.5cm, xshift=2cm] (R) {\bf{Spatial Features}};
\node [block, draw=Violet, below of=R, node distance = 0.8cm, xshift=0cm] (S) {\small NR 2D IQA models};
\node[rectangle,draw=RoyalPurple,line width=0.5mm,dashed, fit=(R) (S)](T) {};
%%%
\node [block2, draw=Plum, line width=0.5mm, below of=K, node distance = 2.5cm, xshift=1.3cm] (U) {\bf{BGGD Model Fit Features}};
\node[rectangle,draw=Brown,line width=0.5mm,dashed, fit=(E) (K) (M) (P) (U)](Q) {};
%%% DMOS scores
\node [block, draw=red, line width=0.5mm, dashed, below of=Q, node distance = 4cm, xshift=-2cm] (W) {\small DMOS scores};
%%%
\node [block3, draw=PineGreen, line width=0.5mm, below of=Q, node distance = 6cm, xshift=0cm] (V) {\bf{Supervised Learning, Regression}};
\path [line1] (A.south)  -- +(0.0em,-0.5em) -- +(-5.66em,-0.5em) -- (B.north);
\path [line1] (A.south)  -- +(0.0em,-0.5em) -- +(5.66em,-0.5em) -- (C.north);
\path [line1] ([xshift=-0cm]I.south)  -- ([xshift=-0cm]J.north);
\path [line1] ([xshift=-0cm]N.south)  -- ([xshift=-0cm]O.north);

%\path [line1] ([xshift=-0cm]K.south)  -- ([xshift=-0cm]U.north);
\path [line1] ([xshift=-0cm]K.south)  -- +(0.0cm,-0.5cm) -- +(1cm,-0.5cm) --(U.north);
\path [line1] ([xshift=-0cm]P.south)  -- +(0.0cm,-0.5cm) -- +(-2.3cm,-0.5cm) --(U.north);

\path [line] ([xshift=-0cm]D.south)  -- ([xshift=-0.178cm]Q.north);
\path [line] ([xshift=-0.3cm]Q.south)  -- ([xshift=-0.3cm]V.north);% temporal to supervised learning
%\path [line] (T.south)  -- ([xshift=1.825cm]V.north);
\path [line] ([xshift=-0cm]D.south)  -- +(0.0em,-1em) -- +(-4.2cm,-1em) -- +(-4.2cm,-6.5cm) -- +(-1.815cm,-6.5cm) --(W.north);% S3D to DMOS
\path [line] ([xshift=0.5cm]W.south)  -- ([xshift=-1.5cm]V.north);% DMOS to supervised learning
\path [line] ([xshift=-0cm]D.south)  -- +(0.0em,-1em) -- +(4.5cm,-1em) -- +(4.5cm,-6.3cm) -- +(2cm,-6.3cm) --(T.north);% S3D to spatial
\path [line] (T.south)  -- ([xshift=1.825cm]V.north);% spatial block to supervised learning

\end{tikzpicture}
\end{center}
\caption{\small Flowchart of the proposed VQUEMODES algorithm.}
\label{fig:flowchart}
\end{figure}
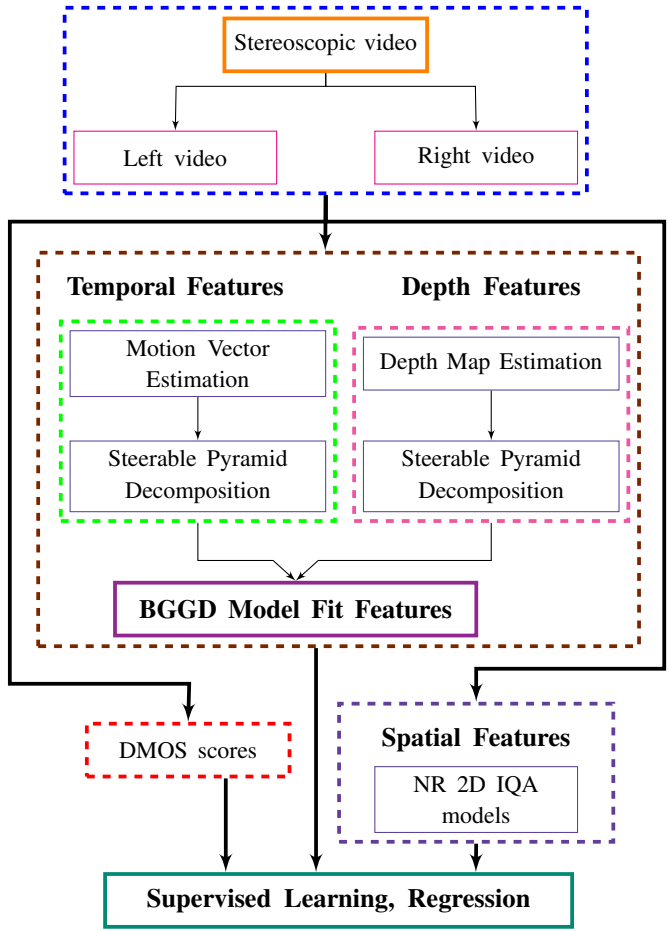

The flowchart of the proposed algorithm is shown in Fig. \ref{fig:flowchart}. The feature extraction stage estimates frame-wise BGGD model parameters to represent motion and depth quality features, and relies on the average 2D NR IQA score as the spatial quality feature. These features are then used to train an SVR for frame-wise quality scores. For video-level quality prediction, the individual frame-wise quality predictions are simply averaged. The algorithm is described in detail in the following. 
\subsection{Feature Extraction}
\subsubsection{Motion Vector Estimation}
The temporal (motion) features are extracted using motion vectors. The motivation to use motion vectors is based on our observations in the previous section that they are as effective as optical flow in distortion discrimination. Also, motion vector computation takes a fraction of the cost of computing the optical flow vectors. We used the three-step search method \cite{jakubowski2013block} to estimate the motion vectors between successive frames and employed a macroblock size of 8 $\times$ 8. Further, the magnitude of the motion vector is used to compute the temporal feature in our algorithm.
\begin{equation*}
M_{s}=\sqrt{M_{H}^{2}+M_{V}^{2}},
\end{equation*}
where, $M_{s}$ represents the motion vector strength, $M_{H}$ and $M_{V}$ are horizontal and vertical motion vector components.
%Fig. \ref{fig:motionvect} shows the motion vector maps of reference and distorted frames from which it is clear that the motion vector statistics are highly disturbed due to the distortions. 
Several video quality models \cite{FLOSIM2015,ha2011perceptual} have effectively used motion vectors as temporal features in their work.
%%%%%%%%%%%%%%%%%%%%%%%%%%%%%%%%%%%%

%%%%%%%%%%%%%%%%%%%%%%%%%%%%%%%%%%%%%%%%%%%%%%%%%%%
%%%%%%%%%%%%%%%%%%%%%%%%%%%%%%%%%%%%%%%%%%%%%%%%%%%%
\subsubsection{Disparity Estimation}
The human visual system takes two retinal views as input and fuses them into a single scene point by converting the scene differences of the two views into disparity/depth information. The simple and complex cells extract the structural properties of both views (spatial and temporal) and further, this information is processed in the primary visual cortex to construct a single scene point illusion of depth perception \cite{levelt1968binocular,hubel1965receptive}. In our work, the disparity maps are computed using an SSIM-based stereo-matching algorithm \cite{chen2013full} for a given stereoscopic frame pair at every time instant. This algorithm works on the principle of finding the best matching block in the right view of a specific block in the corresponding left view. We chose this algorithm based on a trade-off between accuracy and time complexity. Since the temporal features are computed at a block size of $8 \times 8$, we downsampled the disparity map subbands to the same size by averaging over an $8 \times 8$ block.  
\subsubsection{Motion and Depth Feature Extraction}
Once the motion vector map and disparity map is computed for every frame, the BGGD model parameters are estimated using the joint histogram of the subband coefficients of  the motion vector magnitude and the disparity map. As mentioned earlier, we rely on the method in \cite{su2015oriented} for parameter estimation. Specifically, these are computed at three scales and six orientations ($0^0,30^0,60^0,90^0,120^0,150^0$) of the steerable pyramid decomposition. 
\subsubsection{Spatial Feature Extraction}
The third feature we employ in our algorithm is the spatial quality of the video. Spatial quality is computed on a frame-by-frame basis by averaging the spatial qualities of the left and right views of an S3D video. 
\begin{equation*}
S_i=\frac{Spat_{i}^{L}+Spat_{i}^{R}}{2},
\end{equation*}
where, $S_{i}$ represents the overall spatial quality score of the $i^{th}$ frame of an S3D video, $L,R$ represent the left and right views of an S3D video respectively. $Spat$ represents the 2D spatial quality score of the frame as computed using a state-of-the-art 2D NR IQA algorithm.
Specifically, we rely on the following NR IQA algorithms for estimating the spatial quality score of the frames: 
Sparsity Based Image Quality Evaluator (SBIQE) \cite{sbiqepriya}, Blind/referenceless Image Spatial Quality Evaluator (BRISQUE) \cite{mittal2012no}, Natural Image Quality Evaluator (NIQE)  \cite{mittal2013making}.  For brevity, we refer the reader to the respective references for descriptions of these algorithms. All these algorithms deliver state-of-the-art performance on standard IQA databases. 

%%%%%%%%%%%%%%%%%%%%%%%%%%%%%%%%%%%%%%%%%%%
\begin{figure}
\center
\includegraphics[width=8cm]{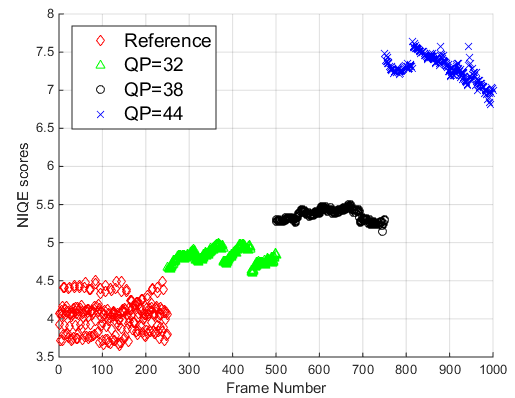}
\caption{Frame-wise NIQE scores of the Boxers video reference sequence and H.264 distortion versions. It is evident that quality variations are clearly tracked.}
\label{fig:featspaNIQE}
\end{figure}
Fig. \ref{fig:featspaNIQE} shows the frame-wise NIQE scores of reference and H.264 distorted versions of the Boxers video sequence of IRCCYN S3D video database. The NIQE scores are clearly varying according to the quality level and is therefore a good discriminatory feature for quality prediction. 
\subsection{Supervised Learning and Quality Estimation}
As mentioned previously, we used three spatial scales and six orientations in our analysis resulting in a total of 18 subbands for every stereoscopic video frame. The BGGD model parameters are computed at every subband resulting in a feature vector $f$= $[\alpha^1\ldots\alpha^{18}; \beta^1\ldots\beta^{18}]$ per frame. For an S3D video, the feature vector set is $[f_1, f_2 \dots f_{n}]$, where $n$ is the number of video frames and $f_{i}$ = $[\alpha_i^1\ldots\alpha_i^{18}; \beta_i^1\ldots\beta_i^{18}]$; $1 \leq i \leq n-1$. The spatial quality feature ($S_{i}$) is appended to the aforementioned BGGD features to explicitly rely on spatial quality. Finally, the feature vector of a video frame is $f_{i}^{s}=[\alpha_i^1\ldots\alpha_i^{18}; \beta_i^1\ldots\beta_i^{18}; S_{i}]$. We believe that over short temporal durations, the average DMOS score of an S3D video and the frame-level DMOS score are highly correlated and are interchangeable. Therefore, we performed the regression of the frame-level features $f_{i}^{s}$ and the video-level DMOS scores $D$ as its label. For video $V$, 
\begin{equation}
f_{i}^{s^{V}}=[\alpha_i^1\ldots\alpha_i^{18}; \beta_i^1\ldots\beta_i^{18};S_{i}],
\label{eqn:consolidated_features}
\end{equation} 
with the corresponding label $D_V$. This feature vector and label set is used to train an SVR. 
SVR is shown to provide good performance even when the available training set size is small, demonstrate accurate performance in one-versus-rest schemes \cite{rifkin2004defense}, provide sparse solutions \cite{scholkopf1997comparing} and accurate estimation of global minimum \cite{cristianini2000introduction} etc. In our work, we used the radial basis function (RBF) kernel as it gave the best overall performance. 

We use regression to estimate the scores of test video frames. It should be noted that the training and regression happen at the frame-level. The overall (video-level) quality score is estimated by averaging the frame-level quality estimates. 

\section{Results and Discussion}
\label{sec:results}
%%%
While several S3D quality assessment databases have been reported in the literature, few of them are open source. We report our results on two popular publicly available databases: the IRCCYN and LFOVIA S3D VQA databases.

The IRCCYN database \cite{urvoy2012nama3ds1} is composed of 10 pristine and 70 distorted video sequences with a good variation in texture, motion and depth components. The video sequences are captured with Panasonic AG-3DA1E twin-lens camera and the baseline separation between lenses is 60 mm. The video sequences are either 16 seconds or 13 seconds in duration and have a resolution of 1920 $\times$ 1080 pixels with a frame rate of 25fps. All the video sequences are encoded in YUV420P format and saved in .avi container. This database is a combination of H.264 and JP2K compression artefact distortions. They utilized the JM reference software to add H.264 compression artefacts by varying the quantization parameter (QP = 32, 38, 44). JP2K artifacts (2, 8, 16, 32 Mb/s) are added on a frame-by-frame basis for both views. These compression artefacts are symmetrically applied on left and right videos. The subjective study is performed in absolute category rating with hidden reference (ACR-HR) method and they published the DMOS values as quality scores. In our evaluation, we limited our temporal extent to the first 10 seconds of the video to avoid issues with blank frames. This applies specifically to the Boxers and Soccer in the database.  

The LFOVIA database \cite{appina2016subjective1} has H.264 compressed stereoscopic video sequences. The database has 6 pristine and 144 distorted video sequences and the videos are encoded in YUV 420P format and saved in mp4 container. The compression artifacts were introduced using \textit{ffmpeg} by changing the bitrate (100, 200, 350, 1200 Kbps) as the quality variation parameter. The video sequences have a resolution of 1836 $\times$ 1056 pixels with a frame rate of 25fps and a duration of 10 sec. This database is a combination of symmetric and asymmetric stereoscopic video sequences. They also performed the subjective study in ACR-HR method and published DMOS values as quality scores. The RMIT database has 47 reference video sequences and does not have any distorted video sequences or subjective scores. Therefore, we did not perform quality assessment on this database.

\begin{table*}[htbp]
\small
\caption{2D and 3D IQA/VQA performance evaluation on IRCCYN and LFOVIA S3D video databases. {\bf{Bold names}} indicate NR QA methods.} % title of Table
\centering % used for centering table
\begin{tabular}{|c| c| c| c| c| c| c|} % centered columns (4 columns)
\hline
  \multirow{2}{*}{\bf Algorithm} & \multicolumn{3}{c|}{\bf IRCCYN Database}& \multicolumn{3}{c|}{\bf LFOVIA Database}\\
 \cline{2-7} 
  &  {\bf LCC}  & {\bf SROCC} & {\bf RMSE} &{\bf LCC}  & {\bf SROCC} & {\bf RMSE} \\
\hline
SSIM \cite{wang2004} &  0.6359	&	0.2465	&	1.0264 &	0.8816 & 0.8828	&	6.1104\\
\hline
MS-SSIM \cite{wang2003multiscale}  &0.9100	&	0.8534	&	0.6512&0.8172	&	0.7888	&	8.9467\\
\hline  
\hline
{\bf{SBIQE}} \cite{sbiqepriya} &0.0081	&	0.0054	&	1.2712& 0.0010	&	0.0043	&	16.0311\\
\hline
{\bf{BRISQUE}} \cite{mittal2012no} &0.7535	&	0.8145	&	0.6535&0.6182	&	0.6000	&	12.6001\\
\hline
{\bf{NIQE}} \cite{mittal2013making} & 0.5729&	0.5664	&0.8464& 0.7206	&	0.7376	&11.1138\\
\hline %   
\hline 
STMAD \cite{vu2011spatiotemporal}& 0.6400 & 0.3495 & 0.9518& 0.6802 & 0.6014 & 9.4918\\
\hline
FLOSIM \cite{FLOSIM2015} &  0.9178&	0.9111	&0.4918 &- &- &-\\
\hline
\hline
Chen \textit{et al.}\cite{chen2013full}  & 0.7886	&	0.7861	&	0.7464&0.8573	&	0.8588	&6.6655\\
\hline
STRIQE \cite{khan2015full} & 0.7931	&	0.6400	&	0.7544& 0.7543	&	0.7485	&	8.5011\\
\hline  
\hline
\bf{VQUEMODES (NIQE)}& \textbf{0.9697}&\textbf{0.9637}&\textbf{0.2635} &\textbf{0.8943}&\textbf{0.8890}&\textbf{5.9124}\\
\hline
\end{tabular}
\label{table:object1}	
\end{table*}
 %%%%%%%%%%%%%%%%%%%%%%%%%%%%%
 \begin{table*}[htbp]
\small
\caption{2D and 3D IQA/VQA performance evaluation on different distortions of IRCCYN S3D video database. {\bf{Bold names}} indicate NR QA methods.} % title of Table
\centering % used for centering table
\begin{tabular}{|c| c| c| c| c| c| c|} % centered columns (4 columns)
\hline
  \multirow{2}{*}{\bf Algorithm} & \multicolumn{3}{c|}{\bf H.264}& \multicolumn{3}{c|}{\bf JP2K}\\
 \cline{2-7} 
  &  {\bf LCC}  & {\bf SROCC} & {\bf RMSE} &{\bf LCC}  & {\bf SROCC} & {\bf RMSE}   \\
\hline
SSIM \cite{wang2004} & 0.7674 & 0.5464 & 0.8843 & 0.7283	&	0.5974	&	0.9202   \\
\hline
MS-SSIM \cite{wang2003multiscale} & 0.8795 & 0.6673 & 0.6955  &0.9414	&	0.9299	&	0.4327 \\
\hline
\hline
{\bf{SBIQE}} \cite{sbiqepriya} & 0.0062 & 0.0058 & 1.9856 & 0.0120	&	0.0574	&	1.0289   \\
\hline % & 
{\bf{BRISQUE}} \cite{mittal2012no} & 0.7915 & 0.7637 & 0.7912  &0.8048	&	0.8999	&	0.5687 \\
\hline 
{\bf{NIQE}} \cite{mittal2013making} & 0.6814 & 0.6412 & 0.8715  & 0.6558	&	0.6427	&	0.7157  \\
\hline
\hline
STMAD \cite{vu2011spatiotemporal}& 0.7641&0.7354 & 0.7296& 0.8388&0.7236 &0.7136 \\
\hline
FLOSIM \cite{FLOSIM2015}& 0.9265 & 0.8987 &0.4256   & 0.9665	&	0.9495	&0.3359\\
\hline
\hline
Chen \textit{et al.} \cite{chen2013full}  & 0.6618 & 0.5720 & 0.6915  & 0.8723	&	0.8724	&	0.6182 \\
\hline
STRIQE \cite{khan2015full} & 0.7430 & 0.7167 & 0.8433  & 0.8403	&	0.8175	&	0.5666  \\
\hline
\hline
%\bf{VQUEMODES with NIQE} &\textbf{0.9762}&\textbf{0.9720}&\textbf{0.5912} &\textbf{0.9816}&\textbf{0.9724}& \textbf{&0.5137} \\
%\hline
\bf{VQUEMODES (NIQE)} &\textbf{0.9594} & \textbf{0.9439} &\textbf{0.1791} &\textbf{0.9859}&\textbf{0.9666}&\textbf{0.0912}\\
\hline
\end{tabular}
\label{table:object2}
\end{table*}
 %%%%%%%%%%%%%%%%%%%%%%%%%%%%%
\begin{table*}[htbp]
\small
\caption{2D and 3D IQA/VQA performance evaluation on symmetric and asymmetric stereoscopic videos of LFOVIA S3D video database. {\bf{Bold names}} indicate NR QA methods.} % title of Table
\centering % used for centering table
\begin{tabular}{|c| c| c| c| c| c| c|} % centered columns (4 columns)
\hline
  \multirow{2}{*}{\bf Algorithm} & \multicolumn{3}{c|}{\bf Symm}& \multicolumn{3}{c|}{\bf Asymm}\\
 \cline{2-7} 
  &  {\bf LCC}  & {\bf SROCC} & {\bf RMSE} &{\bf LCC}  & {\bf SROCC} & {\bf RMSE}   \\
\hline
SSIM \cite{wang2004} & 0.9037 & 0.8991 & 7.0246 & 0.8769 & 0.8755&	5.8162	  \\
\hline
MS-SSIM \cite{wang2003multiscale} & 0.8901 & 0.8681 & 21.2322  &0.8423	&	0.7785	&	15.1681\\
\hline
\hline
{\bf{SBIQE}} \cite{sbiqepriya} & 0.0006 & 0.0027 & 25.4484 & 0.0021	&	0.0051	&	12.0123   \\
\hline
{\bf{BRISQUE}} \cite{mittal2012no} & 0.7829 & 0.7859 & 15.8298  &0.5411	&	0.5303	&	10.1719 \\
\hline 
{\bf{NIQE}} \cite{mittal2013making} & 0.8499 & 0.8705 & 13.4076  & 0.6835	&	0.6929	& 8.8334 \\
\hline
\hline
STMAD \cite{vu2011spatiotemporal}& 0.7815 &0.8000 & 10.2358& 0.6534&0.6010 &9.1614 \\
\hline
\hline
Chen \textit{et al.} \cite{chen2013full}  & 0.9435 & 0.9182 &5.4346 & 0.8370	&	0.8376  & 6.6218\\
\hline
STRIQE \cite{khan2015full} & 0.8275 & 0.8017 &9.2105 & 0.7559	&	0.7492	&7.9321 \\
\hline
\hline
\bf{VQUEMODES (NIQE)} &\textbf{0.9285} &\textbf{0.9236} & \textbf{3.9852}  &\textbf{0.8955}&	\textbf{0.8490}&\textbf{6.9563}  \\
\hline
\end{tabular}
\label{table:object3}
\end{table*}
 %%%%%%%%%%%%%%%%%%%%%%%%%%%%%
%%%%%%%%%%%%%%%%%%%%%%
\begin{table*}
\centering
\caption{Performance evaluation on IRCCYN S3D video database with different 2D NR IQA models in proposed algorithm.}
\begin{tabular}{|c|c|c|c|c|c|c|c|c|c|c|c|}
\hline
&\multirow{2}{*}{\bf Algorithm} & \multicolumn{3}{c|}{\bf H.264}& \multicolumn{3}{c|}{\bf JP2K}& \multicolumn{3}{c|}{\bf Overall}\\
\cline{3-11}
& &  {\bf LCC}  & {\bf SROCC} & {\bf RMSE} &{\bf LCC}  & {\bf SROCC} & {\bf RMSE}  &{\bf LCC}  & {\bf SROCC} & {\bf RMSE} \\
\hline
\multirow{3}{*}{\bf{VQUEMODES}} & SBIQE \cite{sbiqepriya}&0.9262&0.8956&0.3286&0.9706&0.9473&0.2415&0.9622&0.9377&0.3039\\
\cline{2-11}
 & BRISQUE \cite{mittal2012no}  & 0.9517 & 0.9323 & 0.2319 & 0.9809	&	0.9577	&	0.1097 & 0.9624	&	0.9482	&	0.3019  \\
\cline{2-11}
 & NIQE \cite{mittal2013making}  &\textbf{0.9594} & \textbf{0.9439} &\textbf{0.1791} &\textbf{0.9859}&\textbf{0.9666}&\textbf{0.0912}& \textbf{0.9697}&\textbf{0.9637}&\textbf{0.2635}  \\
\cline{1-11}
\hline
\end{tabular}
\label{table:object4}
\end{table*}

%%%%%%%%%%%%%%%%%%%%%%%%%%%%%%%%%%%%%%%%
\begin{table*}
\centering
\caption{Performance evaluation on LFOVIA S3D video database with different 2D NR IQA models in proposed algorithm.}
\begin{tabular}{|c|c|c|c|c|c|c|c|c|c|c|c|}
\hline
&\multirow{2}{*}{\bf Algorithm} & \multicolumn{3}{c|}{\bf Symm}& \multicolumn{3}{c|}{\bf Asymm}& \multicolumn{3}{c|}{\bf Overall}\\
\cline{3-11}
& &  {\bf LCC}  & {\bf SROCC} & {\bf RMSE} &{\bf LCC}  & {\bf SROCC} & {\bf RMSE}  &{\bf LCC}  & {\bf SROCC} & {\bf RMSE} \\
\hline
\multirow{3}{*}{\bf{VQUEMODES}} & SBIQE \cite{sbiqepriya}&0.9000 & 0.8913 & 4.5900 & 0.8532&0.8234&7.1376 & 0.8483&0.8341&7.3476\\
\cline{2-11}
 & BRISQUE \cite{mittal2012no}  & 0.9125 & 0.9013 &4.3980 &0.8792&0.8489&6.8563 &0.8827	&0.8693	&5.9859\\
\cline{2-11}
 & NIQE \cite{mittal2013making}  &\textbf{0.9285} &\textbf{0.9236} & \textbf{3.9852}  &\textbf{0.8955}&	\textbf{0.8490}&\textbf{6.9563}  &\textbf{0.8943}&\textbf{0.8890}&\textbf{5.9124}\\
\cline{1-11}
\hline
\end{tabular}
\label{table:object5}
\end{table*}
%%%%%%%%%%%%%%%%%%%%%%%%%%%%%%%%%%%%%%%%%%%%%%%%%%%%%%%%%%%%%%%%%%%%%%%%%%%%
%%%%%%%%%%%%%%%%%%%%%%%%%%%%%%%%%%%%%%%%%%%%%%%%%%%%%%%%%%%%%%%%%%%%%%%%%%%%%%%%%
\begin{table*}
\caption{Performance comparison with different 3D VQA metrics on IRCCYN S3D video database. {\bf{Bold names}} indicate NR QA methods.}
\centering
\begin{tabular}{|c|c|c|c|c|c|c|c|c|c|}
\hline
\multirow{2}{*}{\bf Algorithm}& \multicolumn{3}{c|}{\bf H.264}& \multicolumn{3}{c|}{\bf JP2K}& \multicolumn{3}{c|}{\bf Overall}\\
\cline{2-10}
&  {\bf LCC}  & {\bf SROCC} & {\bf RMSE} &{\bf LCC}  & {\bf SROCC} & {\bf RMSE}  &{\bf LCC}  & {\bf SROCC} & {\bf RMSE} \\
\hline
Temporal FLOSIM \cite{FLOSIM2015} & 0.6453 & 0.5489 & 0.6958 & 0.8441 & 0.8278 & 0.7027 & 0.7252 & 0.7097 & 0.8528 \\
\hline
{\bf{VQUEMODES}} (no spatial) & 0.9253 & 0.8955 & 0.3555 & 0.9690 & 0.9477 & 0.2572 & 0.9569 & 0.9330 & 0.3162 \\
\hline
PQM \cite{Joveluro2010}& - & - & - & -&-&- & 0.6340&0.6006&0.8784\\
\hline % 
${\text{Chen}}_{3D}$ \cite{Flosim3D2017} & 0.7963 & 0.8035 & 2.5835 & 0.9358 & 0.8884 & 3.2863 & 0.8227 & 0.8201 & 2.9763\\
\hline 
${\text{STRIQE}}_{3D}$ \cite{Flosim3D2017}  & 0.6836 & 0.6263 & 2.3683 & 0.8778 & 0.8513 & 3.2121 & 0.7599 & 0.7525 & 2.8374 \\
\hline
${{\text{FLOSIM}}_{3D}}$ \cite{Flosim3D2017} & 0.9589 & 0.9478 & 0.3863 & 0.9738	&	0.9548	&0.2976 & 0.9178&	0.9111	&0.4918\\
\hline 
PHVS-3D \cite{jin2011frqa3d}& - & - & - & -&-&- & 0.5480&0.5146&0.9501\\
\hline
3D-STS \cite{Han2012VCIP}& - & - & - & -&-&- & 0.6417&0.6214&0.9067\\
\hline
SJND-SVA \cite{qi2016stereoscopic} & 0.5834 & 0.6810 & 0.6672 & 0.8062&0.6901&0.5079 & 0.6503&	0.6229	&0.8629\\
\hline
%SFD \cite{lu2009quality}& - & - & - & -&-&- & 0.5965&0.5896&0.9117\\
%\hline
{\bf{Yang \textit{et al.}}} \cite{yang2017}& - & - & - & -&-&- & 0.8949&0.8552&0.4929\\
\hline
{\bf{BSVQE}} \cite{chen2017blind}& 0.9168 & 0.8857 & - & 0.8953&0.8383&- & 0.9239&0.9086&-\\
\hline
\bf{VQUEMODES (NIQE)} &\textbf{0.9594} & \textbf{0.9439} &\textbf{0.1791} &\textbf{0.9859}&\textbf{0.9666}&\textbf{0.0912}& \textbf{0.9697}&\textbf{0.9637}&\textbf{0.2635}  \\
\hline
\end{tabular}
\label{table:object6}
\end{table*}
%% IRCCYN dataset
\begin{figure}[htbp]
%\centering
\begin{subfigure}[b]{0.24\textwidth}%{0.3\textwidth}
\includegraphics[height=3cm]{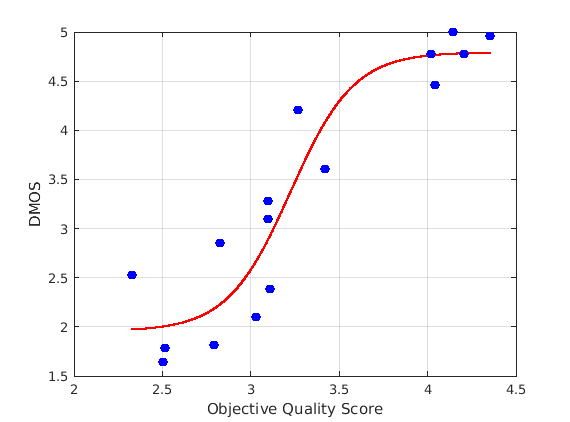}
\subcaption{\small VQUEMODES (no spatial).}
\end{subfigure}
\begin{subfigure}[b]{0.24\textwidth}
\includegraphics[height=3cm]{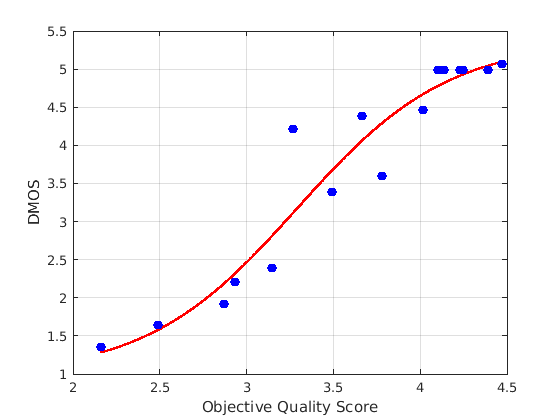}
\subcaption{\small VQUEMODES (SBIQE).} 
\end{subfigure}
\\
\begin{subfigure}[b]{0.24\textwidth}
\includegraphics[height=3cm]{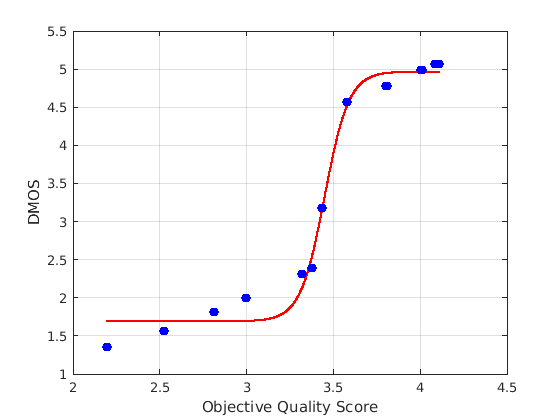}
\subcaption{\small VQUEMODES (BRISQUE).} 
\end{subfigure}
\begin{subfigure}[b]{0.24\textwidth}
\includegraphics[height=3cm]{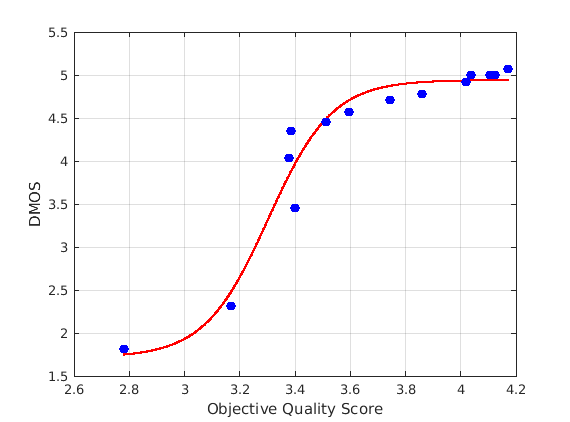}
\subcaption{\small VQUEMODES (NIQE).} 
\end{subfigure}
\caption{Scatter plots of proposed VQUEMODES algorithm prediction without and with different spatial metrics (SBIQE/BRISQUE/NIQE) versus DMOS values on the IRCCYN database.}
\label{fig:scatterIRCCYN}
\end{figure}

%% LFOVIA dataset
\begin{figure}[htbp]
%\centering
\begin{subfigure}[b]{0.24\textwidth}%{0.3\textwidth}
\includegraphics[height=3cm]{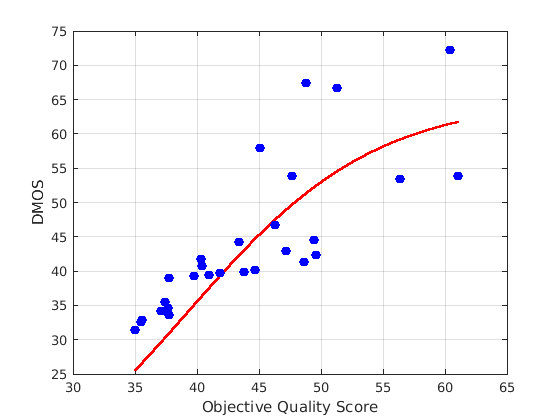}
\subcaption{\small VQUEMODES (no spatial).}
\end{subfigure}
\begin{subfigure}[b]{0.24\textwidth}
\includegraphics[height=3cm]{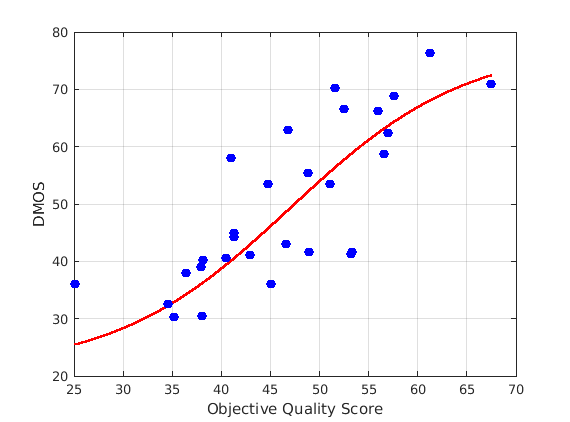}
\subcaption{\small VQUEMODES (SBIQE).} 
\end{subfigure}
\\
\begin{subfigure}[b]{0.24\textwidth}
\includegraphics[height=3cm]{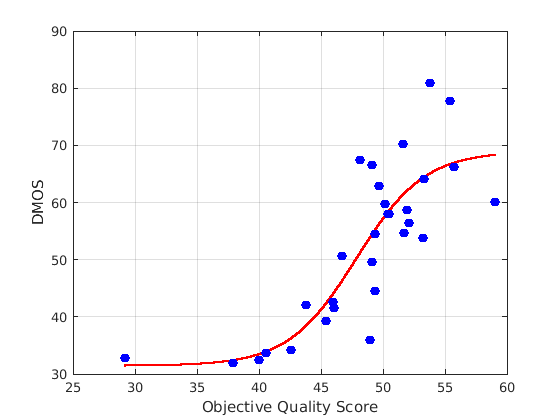}
\subcaption{\small VQUEMODES (BRISQUE).} 
\end{subfigure}
\begin{subfigure}[b]{0.24\textwidth}
\includegraphics[height=3cm]{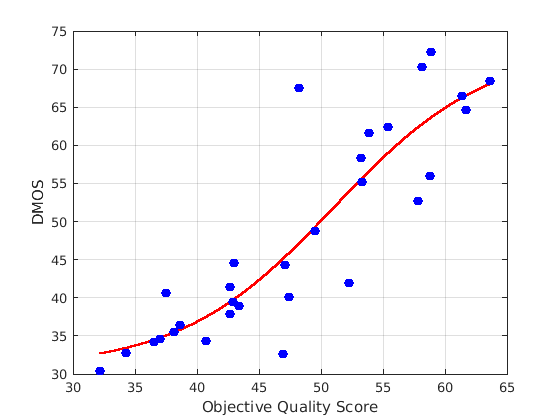}
\subcaption{\small VQUEMODES (NIQE).} 
\end{subfigure}
\caption{Scatter plots of proposed VQUEMODES algorithm prediction without and with different spatial metrics (SBIQE/BRISQUE/NIQE) versus DMOS values on the LFOVIA database.}
\label{fig:scatterLFOVIA}
\end{figure}

For both the databases, 80\% of the videos are used for SVR training and the remaining samples are used for regression. In other words, the training and test sets are obtained by partitioning the set of available videos in the 80:20 proportion. Once this video-level partitioning is done, the actual training happens at the frame-level. During regression, the frame-level scores are estimated and averaged to compute the video-level quality score. We empirically justify the averaging of the frame-level scores to generate the video-level score. In over 1000 regression iterations, we found that the standard deviation of frame-level scores for a given video varied between 0.2 $\times$ $10^{-8}$ and 0.25. This observation, combined with the high correlation of the average scores with DMOS values shown in Tables \ref{table:object1} - \ref{table:object6} provide evidence for the effectiveness of our approach.  
%Table \ref{table:object8} shows the minimum and maximum values of mean ($\mu$),  median ($M$) and standard deviation ($\sigma$) scores of video frame-level regression results over 1000 iterations of IRCCYN dataset.

We used the open-source SVM package {\em{LIBSVM}} \cite{chang2011libsvm} in our experiments. 
%and also, we are motivated from studies \cite{appina2016no,chen2013no} to use the {\em{LIBSVM}}
We performed the training and testing 1000 times for statistical consistency with a random assignment of video-level samples without overlap between the training and testing sets. The reported results are the average over these 1000 trials. The performance of the proposed metric is measured using the following statistical measures: Linear Correlation Coefficient (LCC), Spearman's Rank Order Correlation Coefficient (SROCC) and Root Mean Square Error (RMSE). LCC signifies the linear dependence between two variables and SROCC reveals the monotonic relationship between two quantities. Higher LCC and SROCC value point to a good agreement between the subjective and objective. RMSE quantifies the magnitude of the error between estimated quality scores and DMOS values. All these results are evaluated after performing a non-liner logistic fit. We followed the standard procedure recommended by Video Quality Experts Group (VQEG) \cite{website:LIVE_Database_Report} to perform the non-linear regression with 4-parameter logistic transform given by 
\begin{equation*}
f(x)=\frac{\tau_{1}-\tau_{2}}{1+\text{exp}({\frac{x-\tau_{3}}{|\tau_{4}|}})}+\tau_{2},
\end{equation*}
where {\em{x}} denotes the raw objective score, and $\tau_{1}, \tau_{2}, \tau_{3}$ and $\tau_{4}$ are the free parameters selected to provide the best fit of the predicted scores to the DMOS values. 

Table \ref{table:object1} shows the performance of the proposed metric VQUEMODES (NIQE) on the IRCCYN and LFOVIA databases. We used the NIQE scores as spatial quality score in the proposed algorithm. Also, we compared our metric's performance with different state-of-art 2D IQA/VQA and 3D IQA/VQA models. SSIM \cite{wang2004}, MS-SSIM \cite{wang2003multiscale} are FR 2D IQA metrics and SBIQE \cite{sbiqepriya}, NIQE \cite{mittal2013making} and BRISQUE \cite{mittal2012no} are 2D NR IQA metrics. These IQA metrics were applied on a frame-by-frame basis for each view and the final quality is computed by calculating the average of frame scores of both views. STMAD \cite{vu2011spatiotemporal} and FLOSIM \cite{FLOSIM2015} are 2D FR VQA metrics applied on individual views and the final score is computed by calculating the mean of both view scores. Chen \textit{et al.} \cite{chen2013full} and STRIQE \cite{khan2015full} are stereoscopic FR IQA metrics. These metrics were applied on a frame-by-frame basis for the 3D video and the final quality score is computed by calculating the mean of the frame-level quality scores. From the table it is clear that the proposed metric outperforms all of the 2D IQA/VQA and 3D IQA models.

Table \ref{table:object2} shows the proposed metric's evaluation on different distortions of the IRCCYN database. Table \ref{table:object3} shows the performance evaluation of proposed metric on symmetric and asymmetric S3D video sequences of the LFOVIA database. Symmetric stereoscopic video has both views at same quality and asymmetric stereoscopic video has each view at a different quality. From these results it is clear that the proposed metric has state-of-art performance compared to the other quality metrics in both cases of different distortions (H.264 and JP2K) and symmetric and asymmetric distorted stereoscopic video sequences. 

We checked the efficacy of proposed algorithm by replacing the NIQE scores with other popular 2D NR IQA methods as well. Tables \ref{table:object4} and \ref{table:object5} shows the performance evaluation of proposed metric VQUEMODES on the IRCCYN and LFOVIA databases respectively. From these results it is clear that the proposed metric demonstrates consistent and state-of-art results across spatial metrics and across distortion types. Table \ref{table:object6} shows the performance comparison of the proposed method with different 3D VQA metrics on IRCCYN database. PQM \cite{Joveluro2010}, $FLOSIM_{3D}$ \cite{Flosim3D2017}, PHVS-3D \cite{jin2011frqa3d}, 3D-STS \cite{Han2012VCIP} and SJND-SVQ \cite{qi2016stereoscopic} are S3D FR VQA models. Temporal FLOSIM is a part of the FLOSIM \cite{FLOSIM2015} metric which computes the quality score of a video based on disturbances in motion components. $Chen_{3D}$ and $STRIQE_{3D}$ are 3D FR VQA metrics and these models are extensions of the Chen \textit{et al.} and STRIQE (3D IQA metrics) by including the motion scores computed by Temporal FLOSIM. Yang \textit{et al.} and \text{BSVQE} \cite{chen2017blind} are S3D NR VQA models.

%$FLOSIM_{3D}$ \cite{Flosim3D2017} is a 3D FR VQA metric that computes the quality score of an S3D video based on the relation between motion and depth components.  

To highlight the effectiveness of using joint motion and depth statistics for NR VQA, we report the performance of VQUMODES without including the spatial quality feature in Table \ref{table:object6}. The corresponding scatter plot for this case in shown in Fig. \ref{fig:scatterIRCCYN}. It is clear from these numbers and plots that our proposed statistical features are indeed very effective for S3D NR VQA. It is also clear that 
the combination of joint motion and depth statistical features and spatial quality features (NIQE) has shown a small but consistent performance improvement (compared to the stand-alone motion and depth features) indicating that spatial quality does play a role in S3D VQA. Figs. \ref{fig:scatterIRCCYN} and \ref{fig:scatterLFOVIA} shows the scatter plots of proposed algorithm with different spatial metrics on the IRCCYN and LFOVIA S3D databases respectively. These scatter plots also provide corroborative evidence for the small but important role played by the spatial feature in improving overall NR VQA performance. 
\section{Conclusions and Future work}
\label{sec:conclusions}
Inspired by the neurophysical response of the MT area to motion and depth inputs, we proposed a BGGD model to capture the joint statistical dependencies between motion and disparity subband coefficients of natural S3D videos. 
The utility and efficacy of the proposed BGGD model was demonstrated in an NR stereo VQA application dubbed VQUEMODES. VQUEMODES was evaluated on the IRCCYN, LFOVIA S3D video databases and shown to have state-of-the-art performance compared to the other 2D and 3D IQA/VQA metrics.
We believe that the proposed model could be useful in several applications like depth frame estimation from temporal feature maps, visual navigation, denoising, quality assessment etc. 

\appendices
\ifCLASSOPTIONcaptionsoff
  \newpage
\fi

\bibliographystyle{ieeetr}
\footnotesize
\bibliography{master_ref}

\end{document}